\pdfoutput=1
\documentclass[11pt]{article}

\usepackage{emnlp2021}

\usepackage{times}
\usepackage{latexsym}
\usepackage{graphicx}
\usepackage{amsmath}
\vspace{-2em}

\usepackage[T1]{fontenc}
\usepackage[utf8]{inputenc}

\usepackage{microtype}

\title{KI-BERT: Infusing Knowledge Context for Better Language and Domain Understanding}

\author{Keyur Faldu \thanks{\hspace{0.1cm}Correspondence to k@embibe.com}\\
Embibe \\\And
Amit Sheth \\
University of South Carolina \\\And
Prashant Kikani \thanks{\hspace{0.1cm}Owned implementation and experiments.}\\
Embibe \\\And
Hemang Akabari \\
Embibe\\
}

\begin{document}
\maketitle
\begin{abstract}
Contextualized entity representations learned by state-of-the-art transformer-based language models (TLMs) like BERT, GPT, T5, etc., leverage the attention mechanism to learn the data context from training data corpus. However, these models do not use the knowledge context. Knowledge context can be understood as semantics about entities and their relationship with neighboring entities in knowledge graphs. We propose a novel and effective technique to infuse knowledge context from multiple knowledge graphs for conceptual and ambiguous entities into TLMs during fine-tuning. It projects knowledge graph embeddings in the homogeneous vector-space, introduces new token-types for entities, aligns entity position ids, and a selective attention mechanism. We take BERT as a baseline model and implement the ``Knowledge-Infused BERT'' by infusing knowledge context from ConceptNet and WordNet, which significantly outperforms BERT and other recent knowledge-aware BERT variants like ERNIE, SenseBERT, and BERT\_CS over eight different subtasks of GLUE benchmark. The KI-BERT-base model even significantly outperforms BERT-large for domain-specific tasks like SciTail and academic subsets of QQP, QNLI, and MNLI. 
\end{abstract}

\section{Introduction}
Current advances in deep learning models for natural language processing have been marked by transformer-based language models (TLMs) like BERT \cite{devlin2018bert}, Roberta \cite{liu2019roberta}, T5\cite{raffel2019exploring}, GPT2 \cite{radford2019language}, etc. They have established increasingly superior performance on the array of downstream NLP tasks like sentiment classification, duplicate detection, natural language inference, question answering, etc. These models learn contextualized representation for tokens, entities, and input records, which are used to predict outcomes. Contextualized representations of entities are derived by cleverly mixing the representations learned for vocabulary tokens over the data. These models outperform human baselines on well-defined tasks and datasets under benchmarks like GLUE \cite{wang2018glue} and SuperGLUE \cite{wang2019superglue}. However, these models do not explicitly leverage knowledge context around entities, which becomes a bottleneck when we finetune these models on smaller domain-specific datasets. The research community has made progress in understanding the reasoning capability, its vulnerability, and challenges in acquiring the implicit knowledge by such models \cite{ribeiro2020beyond} \cite{TalmorTCGB20}. 

Knowledge context can be understood as the semantic context of entities and their relationships with neighboring entities in knowledge graphs or ontology. We can motivate example to understand how knowledge context can help for better language and domain understanding. For example, the input record for duplicate detention task \textit{``What would have happened if Facebook were present at the time of World War I?"} and  \textit{``What would have happened if Facebook were present at the time of World War II?"}. Here, \textit{``World War I"} and \textit{``World War II"} are ``conceptual entities" as they carry different and unique conceptual meanings. We could leverage the  knowledge graph ConceptNet\cite{ConceptNet} to derive the conceptual knowledge context. Both these entities would have different neighboring contexts in Concept-Net. Similarly, another example for duplicate detection task is \textit{``What does eat the phone battery quickly"} and \textit{``What would cause the battery on my phone to drain so quickly"}. Words like \textit{``eat"} and \textit{``drain"} are polysemic words and can be classified as ``ambiguous entities", but they carry a similar word sense in this example. Knowledge graph WordNet\cite{Wordnet} gives possible senses for words, with its definition and relationships with similar senses. 

There has been a growing trend of research around the techniques to infuse knowledge from an external knowledge graph to language models either in the pre-training or fine-tuning stage to improve the performance \cite{ERNIE} \cite{kepler} \cite{kbert} \cite{kadapter}. But there is an opportunity to invent a seamless and effective technique to infuse knowledge context for both conceptual and ambiguous entities from different knowledge graphs. We propose a novel technique for “Knowledge Infusion” to infuse knowledge context for conceptual entities and ambiguous entities. While our technique can be applied to any TLMs, for the scope of this paper, we use it over BERT, and hence present “Knowledge-Infused BERT” (KI-BERT).

\begin{figure}[h]
    \centering
    \includegraphics[scale=0.37]{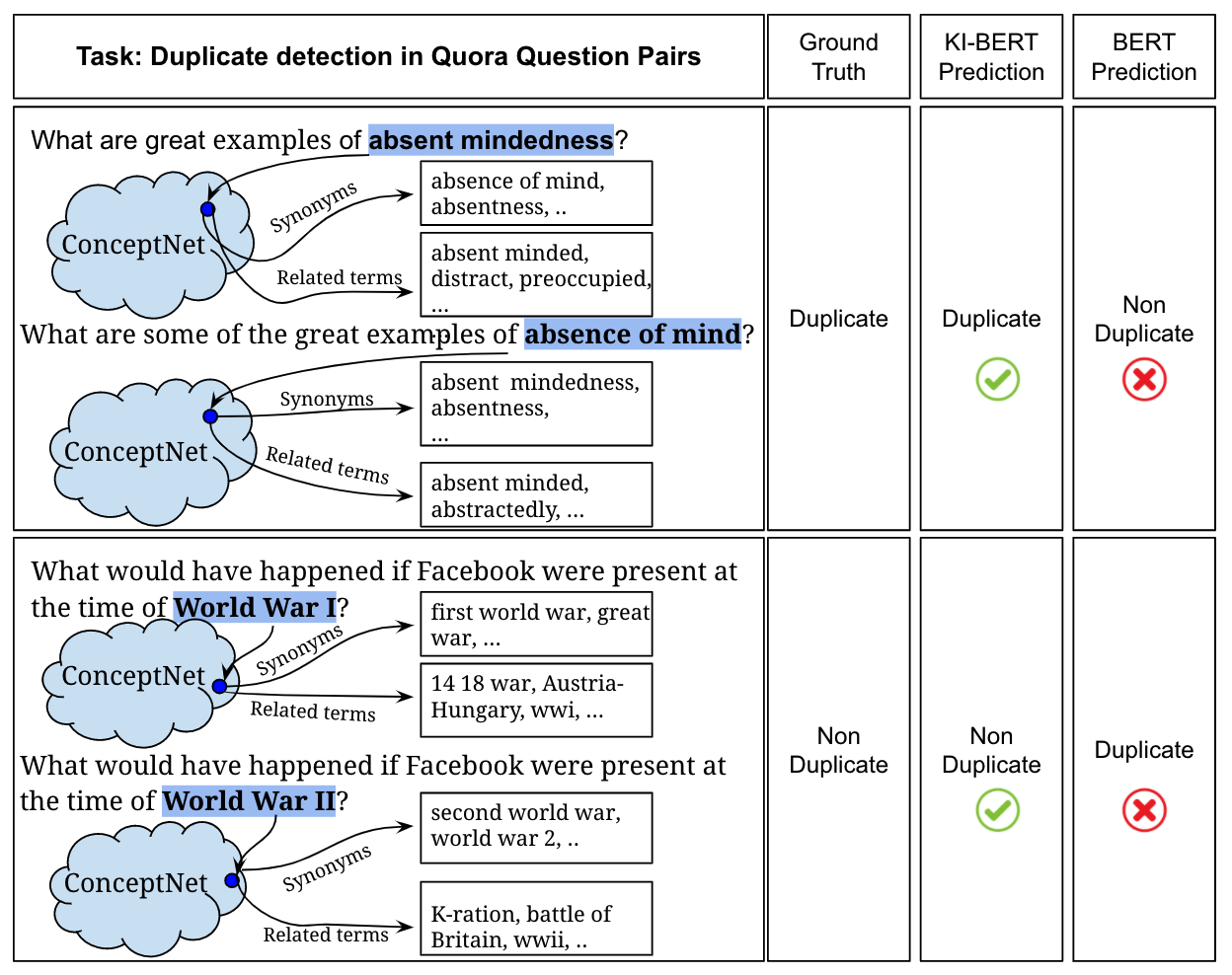}
    \caption{Infusing Knowledge Context for Conceptual Entities}
    \label{fig:mesh1}
\end{figure}

\begin{figure}[h]
    \centering
    \includegraphics[scale=0.33]{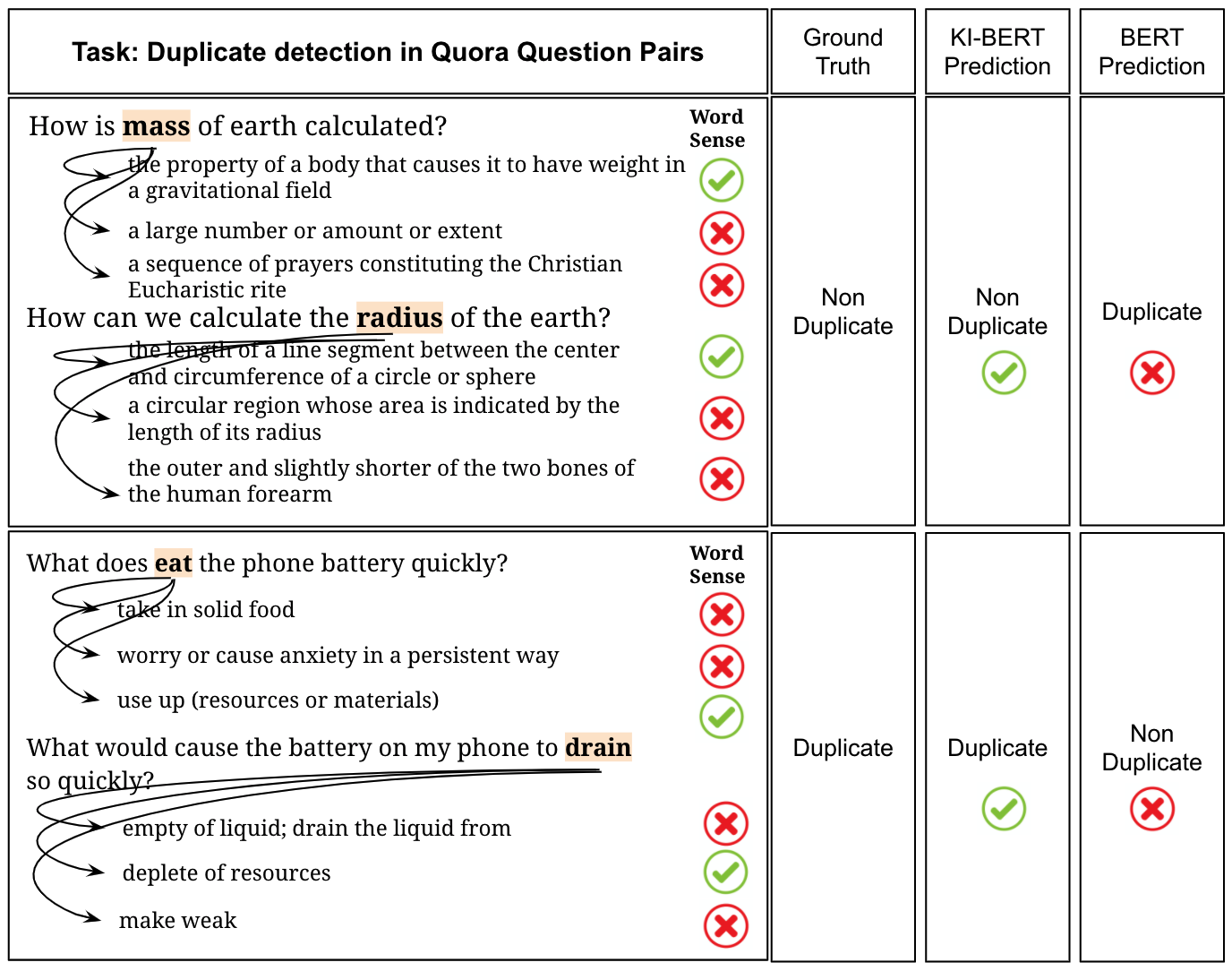}
    \caption{Infusing Knowledge Context for Ambiguous Entities}
    \label{fig:mesh2}
\end{figure}

Figure 1 showcases the examples to infuse knowledge context for conceptual entities from ConceptNet \cite{ConceptNet}. KI-BERT (this work) makes a correct prediction by infusing knowledge context for entities like \textit{``absence of mind"}, \textit{``absent mindedness"}, \textit{``World War I"}, and \textit{``World War II"}. We could notice that BERT fails for these question pairs. Figure 2 illustrates the examples of infusing knowledge context for ambiguous entities from WordNet \cite{Wordnet}. First, we predict the disambiguated sense for ambiguous entities \cite{WSD} like \textit{``mass"}, \textit{``radius"}, \textit{``eat"}, and \textit{``drain"}, and after that knowledge context around disambiguated sense is infused by KI-BERT to make the correct prediction, whereas BERT fails for both these cases.

The key contribution for our work is as follow:

\begin{itemize}
\item{
We propose a novel architecture and technique to infuse external knowledge context during the fine tuning stage into transformer-based language models. We show a specific case of augmenting BERT to derive a novel model ``Knowledge-Infused BERT" (or KI-BERT). We use the entity extraction, entity token types and positional alignment, vector space transformations of knowledge graph embeddings from single or multiple knowledge graphs, and selective attention mechanism to infuse knowledge context. 
 }
\item{
We infuse knowledge context for conceptual entities carrying specific conceptual meaning and ambiguous entities with properties like word polysemy or homophony using knowledge graphs ConceptNet and WordNet, respectively.
}
\item{
 We demonstrate how the KI-BERT significantly improves the performance in comparison to BERT. KI-BERT$_{BASE}$ achieves a performance of 80.47 on GLUE tasks, as compared to 78.9 for BERT$_{BASE}$. KI-BERT$_{BASE}$ outperforms BERT$_{LARGE}$ on domain-specific datasets like SciTail \cite{SciTail}, and domain-specific subsets of QQP, MNLI, and QNLI. Note, BERT$_{LARGE}$ has three times more parameters than KI-BERT$_{BASE}$. We also empirically establish that the performance gain becomes higher as the training data gets smaller, which makes it more suitable to domain tasks with limited labelled data.}
\item{
We plan to release GLUE datasets with tagging of 1) conceptual entities using ConceptNet. 2) ambiguous entities with their predicted word sense (using GlossBERT \cite{GlossBERT}) based on the WordNet-11 dataset\footnote{\label{note1}As of now available on request}}.
\item{
We also plan to release the source code of Knowledge Infused BERT in a git repository\textsuperscript{\ref{note1}}.
}
\end{itemize}

\section{Related Work}

Contextualized representation of an entity is sensitive to the surrounding data context in the training corpus and input record. Contextualized representations learned by TLMs have significantly outperformed feature-based models using non-contextualized embeddings methods like Word2Vec, Glove, etc. TLMs trained on vast corpora derive contextualized representations for entities by cleverly mixing the representation learned for vocabulary tokens over the training data. The task-specific supervised fine-tuning procedure further aligns these contextualized representations  over labeled dataset for better downstream performance.

The knowledge context of an entity can be understood as the semantic context of the entity and its relationships with neighboring entities in knowledge graphs or ontology. Contextualized representations derived using language models need not encode the knowledge context in them. There has been some progress on infusing knowledge from external knowledge graphs into such contextualized representations. 

ERNIE \cite{ERNIE}, is one such popular model that aims to infuse external knowledge for entities using Wikidata by intermixing and projecting back to their respective heterogeneous vector spaces during model pre-training. ERNIE has achieved higher performance in knowledge-driven tasks, however, it could not be seen to improve across most of the GLUE tasks. ERNIE is memory intensive as it effectively doubles the sequence length by reserving entity embeddings for each token.

“Align, mask and select” (AMS) is a method to leverage triples from ConceptNets during the pre-training stage by aligning them with Wikipedia data \cite{Ye_AMS}. It aims to incorporate commonsense knowledge. It gives a good improvement on tasks like CommonSenseQA. But it could not achieve the performance gain over GLUE subtasks. Both ERNIE and AMS approach infuse knowledge during pre-training, which makes them computationally expensive. Whereas KI-BERT infuses during the fine tuning stage makes it effective in low-resource settings. 

K-Adapter \cite{kadapter}, is another method that considers risk of catastrophic forgetting in a pre-trained model, hence, it trains parallel neural adapters for each type of external knowledge during the finetuning. It improves the performance on knowledge-driven tasks like entity typing, relationship classification and NLP tasks like Question Answering using Wikipedia. But it does not report performance on GLUE tasks.

Models like KEPLER \cite{kepler} and K-BERT \cite{kbert} propose directly learning knowledge aware contextualized representations. KEPLER unifies it by incorporating Knowledge Embedding objective along with the Masked Language Modelling objective of BERT. On the other hand, K-BERT introduces a knowledge graph aware soft position and attention visible matrix. KEPLER improves performance for knowledge-driven tasks, but it could not improve performance significantly for GLUE tasks \cite{wang2018glue}. K-BERT improved over domain-specific tasks, but it did not show its performance GLUE tasks. 

All the above-mentioned work was mainly to infuse knowledge context around conceptual entities. On the other hand, ambiguous entities suffer from the problem of word polysemy or homophony. They have different senses in different contexts. Knowledge graph WordNet \cite{Wordnet} lists all possible senses for such ambiguous entities and their relationships with each other. There has been some progress on leveraging knowledge graphs for word senses. EWISE \cite{EWISE} leverages WordNet for the task of Word Sense Disambiguation \cite{WSD}. SenseBERT \cite{SenseBERT} uses a weak supervision method, and uses allowed senses from WordNet to predict the super-sense of the masked word during pre-training. However we could not find any prior work which infuses the knowledge context of disambiguated word sense for ambiguous entities.

Our work demonstrates the infusion of knowledge context for both conceptual entities and ambiguous entities from using a common architecture. Architecture of KI-BERT eases its memory and compute requirements, and it infuses the knowledge during the fine tuning stage, which is computationally less expensive. KI-BERT achieves significant performance gains on GLUE tasks compared to above knowledge aware models.

\section{Knowledge Infused BERT}

KI-BERT extends the architecture of BERT to infuse knowledge context around entities extracted from input records. It projects knowledge graph embeddings of these entities into BERT vector space, introduce new token types for entities, map a positional sequence of these entities, and leverages selective attention to achieve better language and domain understanding capability.

Knowledge embeddings learned over a knowledge graph captures the knowledge context for entities and their relationships with neighbouring entities. We can group these entities into two types 1) conceptual entities, and 2) ambiguous entities. 

1) Conceptual entities: These are the entities representing specific  conceptual meaning. BERT would approximate the semantic representation of such entities by mixing the embedding learned over their sub-words tokens when they are not part of its vocabulary i.e. \textit{“Greenhouse effect”}, \textit{“World War II”}, and \textit{“Refraction”}, etc.  Further, the dataset for the domain-specific task would be dominated by domain-specific entities, and hence there would be more potential to improve performance by infusing knowledge context from relevant knowledge graphs.

2) Ambiguous entities: Entities like frequently used verbs, adjectives, and nouns can suffer from word polysemy or homophony. Where an entity can have different meanings or two different entities can have the same meaning based on the context. For example, the verb \textit{“eat”} can have different meanings, and two different entities \textit{“drain”} and \textit{“eat”} can exactly have the same meaning depending on the context. Knowledge context for such ambitious entities can be derived using two steps. First, predicting the disambiguated sense of ambiguous entities, and then, leveraging the knowledge graph to derive the knowledge context of the disambiguated sense of such ambiguous entities.

Input sequence for KI-BERT is as follow, 
\begin{align*}
([CLS], x_1, x_2, .. x_n, [SEP], x_{n+1}, .., x_m, [SEP], \\
x_{e_1}, x_{e_2}, .., x_{e_{n'}}, [SEP], x_{e_{n'+1}}, .., x_{e_{m'}})
\end{align*} 
Where CLS and SEP are special tokens, $x_1,..x_n$ are tokens from the first sentence and $x_{n+1}, .., x_m$ are tokens from the next sentence. $x_{e_i}$ are entities extracted from the first and second sentence separated by SEP token. 

\begin{figure*}[h]
    \centering
    \includegraphics[scale=0.67]{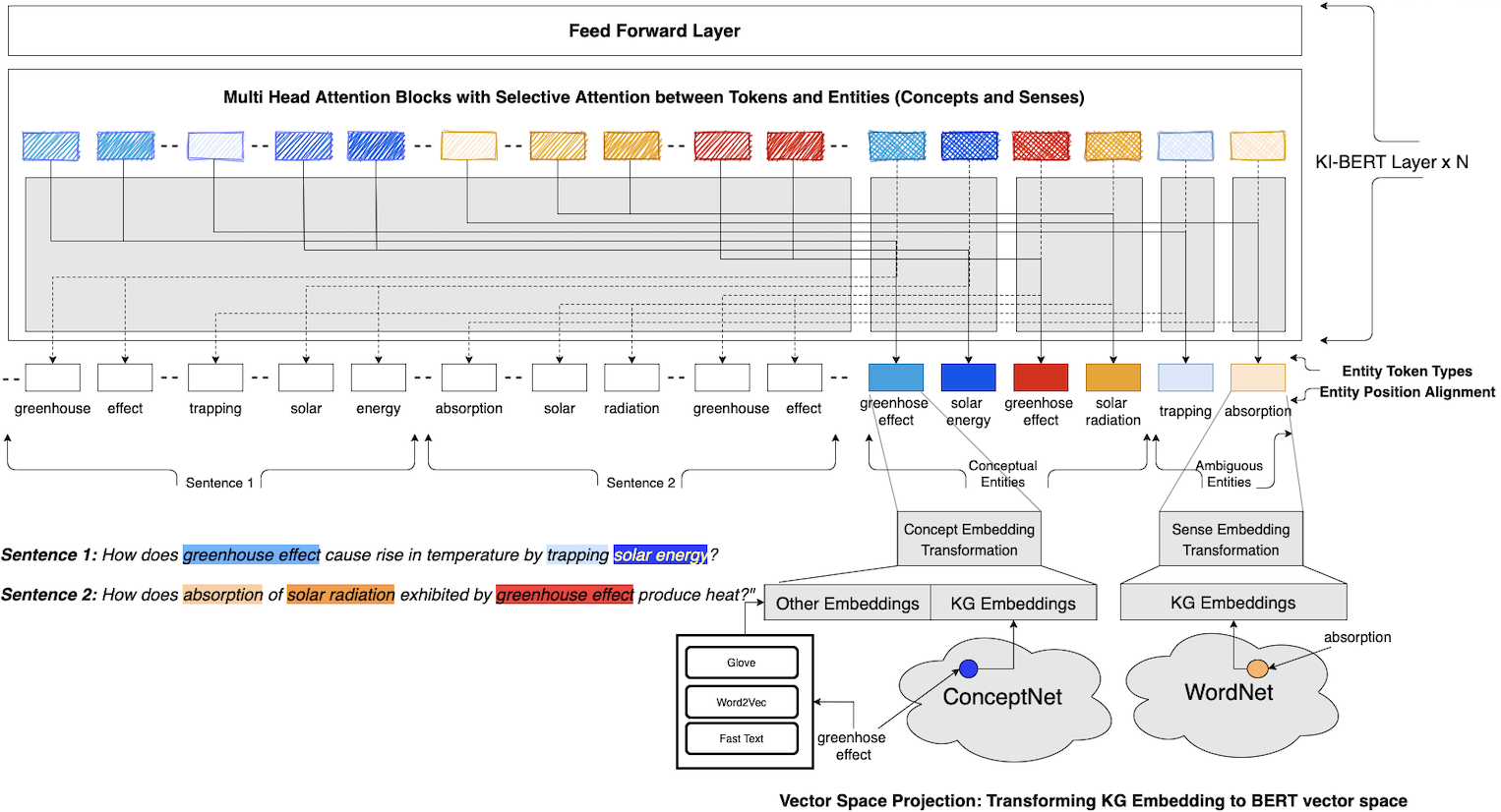}
    \caption{Architecture Diagram of Knowledge Infused BERT. Figure shows a sample input of two sentences 1) \textit{Does the greenhouse effect cause a rise in temperature by trapping solar energy?} and 2) \textit{Does absorption of solar radiation exhibited by the greenhouse effect produce heat?} Conceptual entities (\textit{``greenhouse effect"}, \textit{``solar radiation"}, and \textit{``solar energy"}) and Ambiguous entities (\textit{``trapping"}, \textit{``absorption"}) are extracted. KG embeddings of these entities are then projected to the vector space of BERT. The selective attention mechanism effectively infuse knowledge context from an external source in each layer of KI-BERT.}
    \label{fig:arch}
\end{figure*}

\subsection{Entity Extraction}
Entities are extracted from input data using an explicit n-gram string comparison method with reference entities from the knowledge graph. Once a string is extracted as an entity, we do not extract its substring as another entity. Conceptual entities extraction has an additional constraint that an entity should be an out-of-vocabulary phrase for the model as well. This constraint helps to reduce the number of entities for which knowledge context needs to be added and yet remain effective in overall performance gain. We define a few mapping functions for entities as follow. 

\begin{align*}
EMap(e_i) &= \{x_j, .. x_{j+k}\} \\
\text{where, }&\text{$e_i$ spans over token $x_j,..x_{j+k}$}
\end{align*}
\[
    ESen(e_i)= 
\begin{cases}
    1,& \text{if $e_i$ extracted from sentence 1}\\
    2,              & \text{otherwise}
\end{cases}
\]
\[
    EType(e_i)= 
\begin{cases}
    1,& \text{if $e_i$ is conceptual.}\\
    2,  & \text{if $e_i$ is ambiguous.}
\end{cases}
\]

$EMap$ is a mapping between extracted entities and their corresponding tokens in the input sentence. $ESen$ is a mapping of an entity to its sentence number, and $EType$ maps an entity to its type, which is conceptual or ambiguous. 
\subsection{Vector Space Projection}
Knowledge graph embeddings are a popular method to capture the knowledge context of entities. It either models relationships between entities as mathematical transformations, i.e. transE \cite{TransE}, convE \cite{ConvE}. Or it leverages relationships to aggregate the knowledge context from neighbor entities i.e. GAT \cite{GAT}. KG Embedding gives vector space representation of entities. However, we can not directly inject this representation, as it would confuse the BERT model on how to operate on two heterogeneous vector spaces. This problem becomes more complex when we infuse knowledge context from multiple knowledge graphs. We learn transformations of KG embedding to BERT vector space to effectively leverage knowledge context.
%\begin{align}
%w_{se_i} = TanH(W_{se}^2*Relu(W_{se}^1*wk_{se_i}))\\
%w_{ce_j} = TanH(W_{ce}^2*Relu(W_{ce}^1*wk_{ce_j}))
%\end{align}
\[
w_{e_i}= 
\begin{cases}
    & TanH(W_{C}^2*Relu(W_{C}^1*wk_{e_i})), \\ & \text{\hspace{2cm}if $EType(e_i) = 1$.}\\
    & TanH(W_{A}^2*Relu(W_{A}^1*wk_{e_i})), \\ & \text{\hspace{2cm}if $EType(e_i) = 2$.}\\
\end{cases}
\]
As seen in the equation above, $wk_{e_i}$ is the KG embedding for entity i which is taken from their corresponding knowledge graphs depending upon if $e_i$ is conceptual or ambiguous. We build a two layer feed forward network, where weights $W_{C}^1$ and $W_{C}^2$ (similarly $W_{A}^1$ and $W_{A}^2$) are learnable weights for transforming conceptual entities (ambiguous entities). 

We have found empirically that vector space transformations are more effective when higher dimensional embeddings are projected into lower dimensional space. If KG embeddings have lower dimensionality, we can concatenate it with other embeddings derived independently, like Word2Vec, Glove, or FastText. Such external representations could help infusing additional semantic information about entities and deriving better transformations of its representations to BERT vector space.

\subsection{Entity Token Types}
 
BERT has two different token type ids, tokens from the first segment/sentence are given token-type $0$, and tokens from the second segment/sentence are given token-type $1$. Since KI-BERT  additionally feeds extracted entities from the first sentence and second sentences, it assigns them token-type $2$. 

\subsection{Entity Position Alignment}

BERT mixes input embeddings of tokens with the embeddings corresponding to the positional sequence before feeding it to the bottom-most layer. This helps BERT learn the linguistic capabilities like POS tagging, dependency relations, semantic role labeling, and co-reference resolution, etc \cite{BERTlookat}. It would be not meaningful to assign an increasing position sequence for entities, as these entities are extracted from tokens present at different positions in the input record. KI-BERT assigns position ids for entities based on the position of corresponding tokens.

\begin{table*}[hbt!]
\fontsize{10pt}{15pt}
\selectfont
\begin{tabular}{lccccccccc}
\hline System & MNLI-(m/mm) & QQP & QNLI & SST-2 & CoLA & STS-B & MRPC & RTE & Average \\
& 392k & $363 \mathrm{k}$ & $108 \mathrm{k}$ & $67 \mathrm{k}$ & $8.5 \mathrm{k}$ & $5.7 \mathrm{k}$ & $3.5 \mathrm{k}$ & $2.5 \mathrm{k}$ & $-$ \\
\hline BERT $_{\text {BASE }}$ & $84.6 / 83.4$ & 71.2 & 90.5 & 93.5 & 52.1 & 85.8 & 88.9 & 66.4 & 79.6 \\
%BERT $_{\text {LARGE }}$ & $86.7/85.9$ & 72.1 & 92.7 & 94.9 & 60.5 & 86.5 & 89.3 & 70.1 & 82.1 \\
SenseBERT & $83.6 / 83.4$ & 70.3 & 90.6 & 92.2 & 54.6 & 83.5 & 89.2 & 67.5 & 79.43 \\
ERNIE & $84.0/83.2$ & 71.2 & 91.3 & 93.5 & 52.3 & 83.2 & 88.2 & 68.8 & 79.5 \\
BERT CSbase & $84.7/83.9$ & 72.1 & 91.2 & 93.6 & 54.3 & 86.4 & 85.9 & 69.5 & 80.17 \\
\hline BERT $_{\text {BASE }}$ (Ours) & $83.9 / 83$ & 70.1 & 90.9 & 93.3 & 51.7 & 84.3 & 88.3 & 64.6 & 78.9 \\
KI-BERT-Sense $_{\text {BASE }}$ & $84.6 / 83.4$ & 71.4 & \textbf{91.3} & 93.6 & \textbf{55.8} & 83 & \textbf{88.5} & 69.1 & 80.17 \\
KI-BERT-Concepts $_{\text {BASE }}$ & \textbf{$85.2 / 83.7$} & \textbf{71.5} & 91.2 & \textbf{94.4} & 55.5 & \textbf{85.1} & 88.2 & \textbf{69.3} & \textbf{80.46} \\
\hline
\end{tabular}
\caption{Comparing KI-BERT performance with other models for GLUE tasks}
\label{tab:expglue}
\end{table*}

\subsection{Selective Attention}
Multi-head attention layer in transformer architecture enables each token in the input record to attend to all other tokens. Entities are appended after the tokens from the input record, with projected dense vector representations from KG-embeddings of knowledge graphs. These entities are extracted from the input record and each entity spans over one or many consecutive tokens in the input record. KI-BERT uses ``selective attention", where an entity can attend only those tokens which it spans across, and vice-versa. Further, in the later section, the ablation study also demonstrates the effectiveness of such a mechanism. Entities belonging to the same sentence in an input record also attend to each other. It would possibly help KI-BERT to derive contextualized knowledge context for entities. By not allowing every token to attend all entities and vice-versa, KI-BERT prevents aggressive information flow between tokens and entities. Thus, selective attention streamlines the information flow between entities and tokens, where data context and knowledge context get intermixed to derive a better representation of entities. \\
All the tokens would attend to other tokens, exactly like BERT. Selective Attention between tokens are defined as follow:
\begin{align*}
SA(x_i, x_j) &= 1 \text{, where, $i,j \in \{1,..m\}$}
\end{align*}
%Selective Attention between tokens and specific entities:
%\begin{align*}
%&SA(x_i, x_{se_j}) = 1 \text{, if $x_i \in EntityMap(se_j)$} \\
%&SA(x_{se_i}, x_{se_j}) = 1, \\
%&\hspace{2 cm}\text{ if $se_i$, $se_j$ belongs to same sentence} \\ 
%&SA(x_{se_i} x_{se_i}) = 1 \\
%\end{align*}
%Selective Attention between tokens and common entities:
%\begin{align*}
%&SA(x_i, x_{ce_j}) = 1 \text{, if $x_i \in EntityMap(ce_j)$} \\
%&SA(x_{ce_i}, x_{ce_j}) = 1, \\
%&\hspace{2 cm}\text{ if $ce_i$, $ce_j$ belongs to same sentence} \\
%&SA(x_{ce_i}, x_{ce_i}) = 1
%\end{align*}
%Selective Attention for entities is defined as follow:
%\begin{align*}
%&SA(x_i, x_{e_j}) = 1 \text{, if $x_i \in EMap(e_j)$} \\
%&SA(x_{e_i}, x_{e_j}) = 1, \\ 
%&\text{\hspace{2 cm} if $ESen(e_i) = ESen(e_j)$ and }\\
%&\text{\hspace{2 cm} $EType(e_i) = EType(e_j)$.} 
%\end{align*}
Selective Attention between tokens and entities are defined as follow:
\[
SA(x_i, x_{e_j}) = 
\begin{cases}
& 1 \text{ \hspace{0.2cm} if $x_i \in EMap(e_j)$}\\
& 0 \text{\hspace{0.2cm} otherwise}.\\
\end{cases}
\]
\[
SA(x_{e_j}, x_i) = SA(x_i, x_{e_j})\text{\hspace{2.4cm}}
\]
Selective Attention between entities are defined as follow:
\[
SA(x_{e_i}, x_{e_j}) =
\begin{cases}
& 1 \text{\hspace{0.2cm}if $ESen(e_i) = ESen(e_j)$} \\ & \text{\hspace{0.2cm} and $EType(e_i) = EType(e_j)$}\\
& 0 \text{\hspace{0.2cm} otherwise.}\\
\end{cases}
\]

%Selective attention for all other pairs of entities and tokens, or entities and entities would be marked zero. 
KI-BERT would mask attention between a pair of a token $x_i$ and entity $x_{ej}$ when $SA(x_i, x_{ej}) = 0$.
\begin{table*}
\fontsize{10pt}{15pt}
\selectfont
\begin{tabular}{lccccc}
\hline System & SciTail & QQP(Academic) & QNLI (Academic) & MNLI (Academic) & Average\\
\hline
BERT $_{\text {BASE }}$ (Ours) & 90.97 & 71.94 &  81.64 & 61.36 & 76.47 \\
BERT $_{\text {LARGE }}$ (Ours) & 92.89 & 74.79 & 84.17 & 65.15 & 79.25 \\
KI-BERT-Concepts $_{\text {BASE }}$ & 92.89 & 77.46 & 87.34 & 64.39 & 80.83\\
KI-BERT-ConSen $_{\text {BASE }}$ & \textbf{93.55} & \textbf{77.51} & \textbf{87.56} & \textbf{69.7} & \textbf{82.08}\\
\hline
\end{tabular}
\caption{Comparing KI-BERT performance for domain specific tasks}
\label{tab:expdomain}
\end{table*}

\begin{table*}
\fontsize{10pt}{15pt}
\selectfont
\begin{tabular}{lccccccccc}
\hline System & Parameters & SciTail (15\%) & SciTail (30\%) & SciTail (50\%) & SciTail (100\%)\\
\hline
BERT $_{\text {BASE }}$ (Ours) & 110M & 85.74 & 87.44 & 90.22 & 90.97 \\
BERT $_{\text {LARGE }}$ (Ours) & 330M & 90.26 & 91.76 & 91.25 & \textbf{92.89} \\
KI-BERT-Concepts $_{\text {BASE }}$ & 111M & \textbf{90.82} & \textbf{92.28} & \textbf{92.05} & \textbf{92.89} \\
\hline
\end{tabular}
\caption{Model comparison on different size of training data for SciTail}
\label{tab:expscitail}
\end{table*}

\begin{table}[hbt!]
\fontsize{9pt}{14pt}
\selectfont
\begin{tabular}{lc}
\hline System & GLUE Average \\
\hline
KI-BERT-Concepts $_{\text {BASE }}$ & 80.46 \\
KI-BERT-Concepts $_{\text {BASE }}$ - SA & 79.51\\
KI-BERT-Concepts $_{\text {BASE }}$ - ETT - PA & 79.32 \\
KI-BERT-Concepts $_{\text {BASE }}$ - VSP + PCA & 79.67 \\
\hline
\end{tabular}
\caption{Ablation study of KI-BERT-Concepts $_{\text{BASE}}$ model over GLUE tasks}
\label{tab:expablation}
\end{table}

\section{Experiments}
General Language Understanding Evaluation (GLUE)\cite{wang2018glue} is a multi-task benchmark for natural language understanding. It is a collection of NLU tasks including question answering, sentiment analysis,  and textual entailment. It also has an associated online platform for model evaluation, comparison, and analysis. We have used the GLUE benchmark to compare the performance of Knowledge Infused BERT. 

We have used ConceptNet \cite{ConceptNet} to infuse knowledge context for conceptual entities, and WordNet \cite{Wordnet} for the disambiguated sense of the ambiguous entities. KG Embeddings for ConceptNet are trained with the TransE \cite{TransE} method. We use GlossBERT \cite{GlossBERT} to predict the sense of the ambiguous entities like verbs, nouns, adjectives, etc. We have used KG embeddings for word sense trained with ConvE \cite{ConvE} method over the WordNet11 dataset. 

We have derived three variants of KI-BERT. KI-BERT-Concepts uses infused knowledge context for conceptual entities, KI-BERT-Sense uses infused knowledge context for ambiguous entities and KI-BERT-ConSen uses infused knowledge context for both conceptual and ambiguous entities. We ran all over experiments on V100 GPU with batch size 16. We reproduced BERT$_{BASE}$ with batch size 16, and reported it as BERT$_{BASE}$ (Ours).

\subsection{Language Understanding}

We have compared KI-BERT-Sense$_{BASE}$ and KI-BERT-Concepts$_{BASE}$ models with BERT$_{BASE}$, and knowledge-aware models like ERNIE\cite{ERNIE}, SenseBERT\cite{SenseBERT} and BERT CSbase\cite{Ye_AMS} over General Language Understanding Evaluation (GLUE) benchmark. ERNIE infuses the external knowledge for entities during the pre-training process. SenseBERT supplies additional information like the super sense of input words, where possible, and aims to predict the masked whole word with its super sense during the pre-training process. BERT CSbase model leverages knowledge graphs in its pre-training process to infuse knowledge context. 

As we can notice in Table \ref{tab:expglue}, both KI-BERT-Sense and KI-BERT-Concepts have improved considerably to BERT, ERNIE, SenseBERT, and BERT CSbase over all the GLUE tasks. KI-BERT-Concepts$_{BASE}$ have achieved the average GLUE score of 80.46 across eight different GLUE tasks, which in comparison to the BERT$_{BASE}$ (ours) is higher by 1.56.

\subsection{Domain Understanding}
We have used KI-BERT-ConSen and KI-BERT-Concepts for domain-specific tasks. There are a wide range of problems in the EduTech domain like academic questions de-duplication, question answering, content search, score improvement etc \cite{faldu2020system} \cite{thomas2020system} \cite{faldu2020adaptive}. These problems have the opportunity to infuse knowledge context about domain-specific conceptual entities, which would not be part of BERT vocabulary. We have experimented on natural language inference and question similarity tasks. We have selected academic domain dataset SciTail \cite{SciTail} and derived academic subsets of GLUE tasks like MNLI, QNLI and QQP. Subsets are derived if extracted entities are part of the academic knowledge graph at (redacted). As observed in Table \ref{tab:expdomain}, KI-BERT-ConSen$_{BASE}$ significantly outperforms both BERT$_{BASE}$ and BERT$_{LARGE}$ over all these domain-specific datasets.

\section{Ablation Study and Analysis}

\subsection{Limited Labelled Domain Datasets}

The availability of labeled data for domain-specific tasks is a big challenge, so, we have analyzed how KI-BERT performs when we only train it on a subset of training data. As we can observe in Table \ref{tab:expscitail}, the relative improvement from KI-BERT-Concepts becomes increasingly higher as the size of training data decreases. KI-BERT-Concepts$_{BASE}$ having 111 million parameters outperforms BERT$_{LARGE}$ which has 330 million parameters. Moreover, KI-BERT infuses knowledge during finetuning stage, which also makes it easy to adapt it to various domain specific tasks.

\subsection{Contribution of each Novel Techniques}
We have performed ablation analysis to understand how each novel decision choice made for KI-BERT is adding value. Table \ref{tab:expdomain} mentions the average performance over GLUE tasks for different variants of KI-BERT-Concepts$_{BASE}$.  $\text{KI-BERT-Concepts}_{BASE}\text{ - SA}$ is a model variant which does not use the selective attention mechanism explained in section 3.5. Similarly, $\text{KI-BERT-Concepts}_{BASE}\text{ - ETT - PA}$ variant does not use special entity token types 2 defined in section 3.3, and also, it does not have a positional alignment for entities defined in section 3.4. The model variant $\text{KI-BERT-Concepts}_{BASE}\text{ - VSP}$ does not use vector space projection and instead uses principle component analysis to get top 768 (i.e. dimensions for BERT vector space) dimensions from the vector spaces of knowledge graph embedding. As we could notice in Table \ref{tab:expablation}, the performance of KI-BERT drops if we remove any of the novel techniques.

\begin{figure}[h]
    \centering
    \includegraphics[scale=0.17]{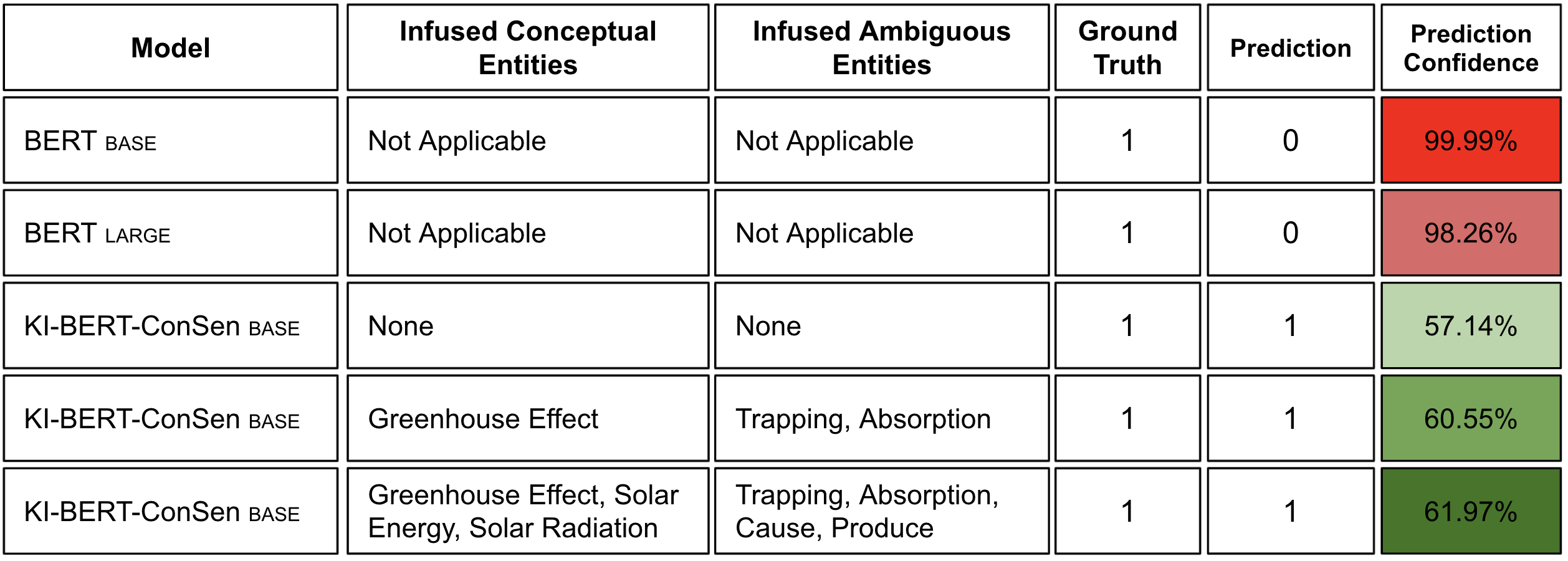}
    \caption{Analysis of KI-BERT-ConSen prediction confidence for duplicate detection for two questions (i) \textit{Does greenhouse effect cause rise in temperature by trapping solar energy?} (ii) \textit{Does absorption of solar radiation exhibited by greenhouse effect produce heat?}}
    \label{fig:analysis}
\end{figure}

\subsection{How Infusing Knowledge Context Helps?}
As shown in Figure \ref{fig:analysis}, we infer on a given example using KI-BERT-ConSen, it makes the right prediction even when we don't explicitly infuse knowledge context around entities. This could be because KI-BERT would have already acquired knowledge context while training data in its model parameters. On the contrary, both BERT-base and BERT-large models make a wrong prediction with very high confidence of 99.99\% and 98.27\%, respectively. Confidence of correct prediction improves by 3.3\% by infusing knowledge context of entities \textit{``greenhouse effect"}, \textit{``absorption"}, and \textit{``trapping"}. It further improves by 1.5\% when we infuse knowledge context of other entities \textit{``solar radiation"}, \textit{``solar energy"}, \textit{``produce"} and \textit{``cause"}. 

\section{Conclusion and Future Work}

We have proposed a novel technique to infuse the knowledge context of entities for better semantic capabilities of state-of-the-art TLMs. We have categorized entities into conceptual and ambiguous entities. Ambiguous entities like verbs, nouns, adjectives are ambiguous in nature and suffer from the problem of word polysemy or homophony. On the other hand, conceptual entities have specific meanings, but they may not be present in the vocabulary of language models. We proposed, implemented, and validated novel techniques like vector space projection, entity token types, entity position alignment, and selective attention, which could effectively infuse knowledge context from knowledge graphs into transformer-based language models. We took BERT as a specific use case to infuse knowledge context and releases KI-BERT (Knowledge Infused BERT) which outperforms BERT, and other knowledge augmented models like ERNIE \cite{ERNIE}, SenseBERT \cite{SenseBERT}, BERT CSbase \cite{Ye_AMS} etc. Further, KI-BERT$_{BASE}$ significantly outperforms BERT$_{LARGE}$ on domain-specific tasks, which has three times higher parameters. KI-BERT is suitable for tasks with limited labelled dataset, as it significantly outperforms BERT in such scenarios.

In this work, we have shown how to infuse conceptual and lexical knowledge from ConceptNet and WordNet. We plan to further extend the best way to infuse other forms of knowledge context like syntactic knowledge, domain knowledge, procedural knowledge etc. KI-BERT infuses knowledge context during fine-tuning. Also, currently, KI-BERT infuses knowledge at the bottom-most layer and afterward, it propagates further. We could investigate how we could optimally infuse different types of knowledge (i.e. syntactic, linguistic, conceptual, procedural etc) from different layers. Also, we could explore a multi-task learning setup where the objective function could be based on both downstream task performance and knowledge infusion effectiveness. On the other hand knowledge infusion also has the potential to improve the interpretability and explainability \cite{SemanticsBB} \cite{gaur2020explainable}, which we plan to empirically establish in future. It would be also interesting to take the concept of domain specific knowledge infusion beyond NLP problems like score improvement \cite{donda2020framework} and student behavioral intervention \cite{faldu2020behavioral}.

\section{Acknowledgements}
The authors express their gratitude to Aditi Avasthi, CEO, Embibe for continuous support and direction, and to the Data Science team for helping us during the experimentation.

\bibliography{anthology,custom}
\bibliographystyle{acl_natbib}

\appendix

\end{document}